\documentclass[final]{cvpr}
\usepackage{multirow}
\usepackage[dvipsnames]{xcolor}
\usepackage{times}
\usepackage{booktabs}
\usepackage{makecell}
\usepackage{epsfig}
\usepackage{graphicx}
\usepackage{amsmath}
\usepackage{amssymb}
\newcommand*{\Scale}[2][4]{\scalebox{#1}{$#2$}}%

\usepackage[pagebackref=true,breaklinks=true,colorlinks,bookmarks=false]{hyperref}

\def\cvprPaperID{4306} 
\def\confYear{CVPR 2021}

\begin{document}

\title{BiconNet: An Edge-preserved Connectivity-based Approach for Salient Object Detection}

\author{Ziyun Yang$^{1}$\quad Somayyeh Soltanian-Zadeh$^{1}$ \quad Sina Farsiu$^{  1,2,3,4}$\\
$^{1}$Department of Biomedical Engineering, Duke University, USA\\
$^{2}$Department of Ophthalmology, Duke University Medical Center, USA\\
$^{3}$Department of Electrical and Computer Engineering, Duke University, USA\\
$^{4}$Department of Computer Science, Duke University, USA\\

}

\maketitle

\begin{abstract}
   Salient object detection (SOD) is viewed as a pixel-wise saliency modeling task by traditional deep learning-based methods. A limitation of current SOD models is insufficient utilization of inter-pixel information, which usually results in imperfect segmentation near edge regions and low spatial coherence. As we demonstrate, using a saliency mask as the only label is suboptimal. To address this limitation, we propose a connectivity-based approach called bilateral connectivity network (BiconNet), which uses connectivity masks together with saliency masks as labels for effective modeling of inter-pixel relationships and object saliency. Moreover, we propose a bilateral voting module to enhance the output connectivity map, and a novel edge feature enhancement method that efficiently utilizes edge-specific features. Through comprehensive experiments on five benchmark datasets, we demonstrate that our proposed method can be plugged into any existing state-of-the-art saliency-based SOD framework to improve its performance with negligible parameter increase.
\end{abstract}

\section{Introduction}

\begin{figure}[h!]
\begin{center}
\setlength{\abovecaptionskip}{0.cm}
\includegraphics[width=1\linewidth]{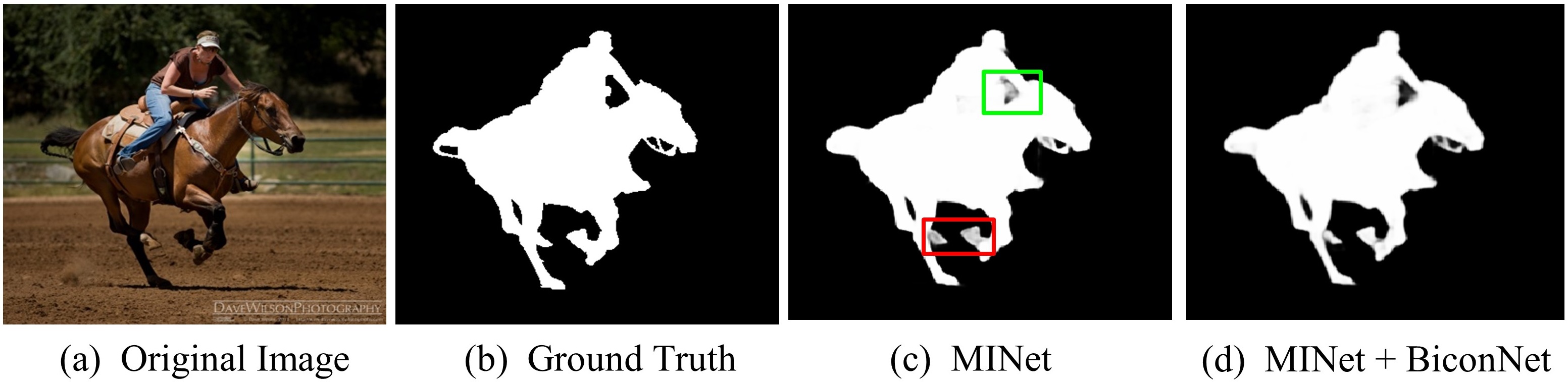}
\end{center}
\vspace{-9pt}
\caption{An example of insufficient modeling of pixel-wise relationship and structural information. MINet \cite{MINet} results in both blurred edges ({\color{green}green} box) and spatial inconsistency problems ({\color{red}red} box). However, our model (MiNet + BiconNet) results in sharper edges and uniformly highlighted predictions near the boundaries.}
\vspace{-10pt}
\label{insufficient}
\end{figure}

As a fundamental task in computer vision, salient object detection (SOD) plays an essential role in image scene understanding \cite{review} and has been applied to different tasks, such as weakly supervised semantic segmentation \cite{weaksup1,weak_PR}, visual tracking \cite{visual_track_PR}, scene analysis \cite{scene1,scene2}, video processing \cite{video_PR} and medical image analysis \cite{medical1}. Convolutional neural networks (CNNs) have greatly promoted the development of SOD due to their capacity to extract multi-level semantic information. Most current CNN-based SOD models \cite{DSS,U2-Net} view the problem as a pixel-level saliency classification task; i.e., their only goal is to assign a saliency score to individual pixels. Despite promising results, these models are limited by insufficient utilization of edge information, and insufficient attention to inter-pixel relationships. These problems together can result in blurred edges or low spatial coherence (i.e., have inconsistent saliency predictions for neighboring pixels that share similar spatial features), as Fig. \ref{insufficient}.

\begin{figure*}[h!]
\begin{center}
\setlength{\abovecaptionskip}{0.cm}
\includegraphics[width=0.98\linewidth]{{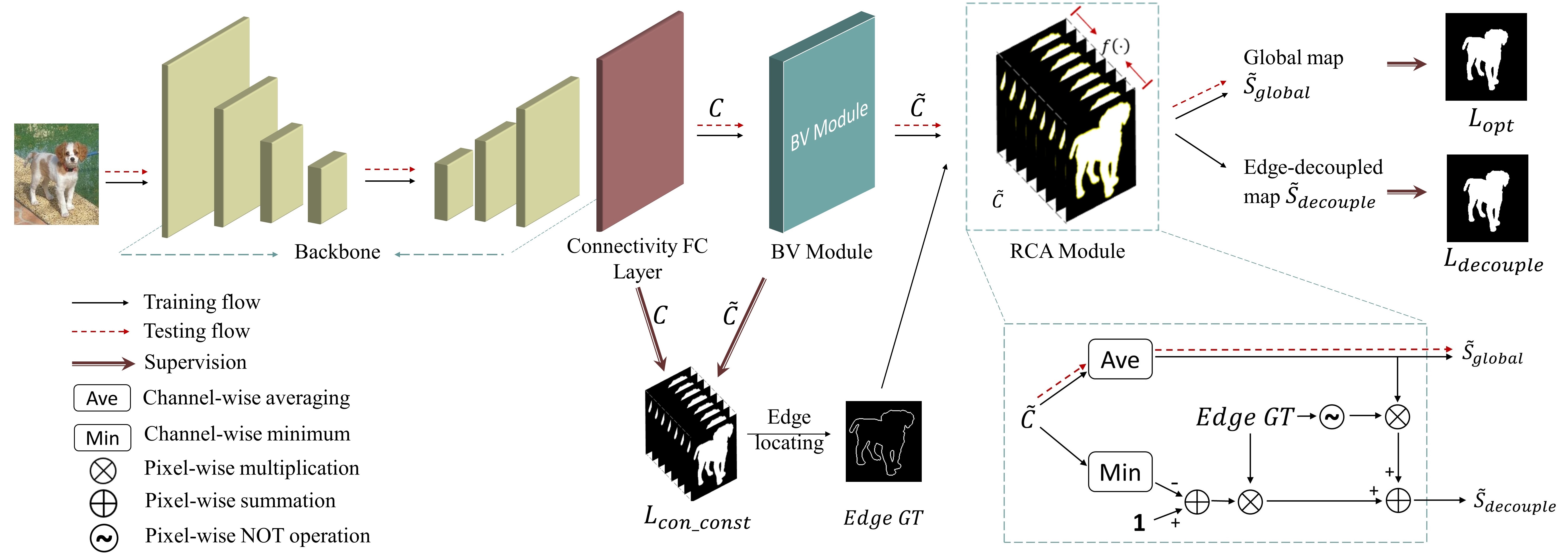}}
\end{center}
\vspace{-9pt}
   \caption{The overview of BiconNet, which contains a backbone, an 8-channel connectivity fully connected layer, a BV module, and an RCA module. Note that we can directly get edge information from the ground truth connectivity map and use it for highlighting the edge-specific features in the RCA module.}
\label{overvew}
\vspace{-8pt}
\end{figure*}

The edge problem has been alleviated somewhat by adding edge information into networks using extra supervision flows \cite{egnet,scrn,ITSD}, but there is still room for impactful improvement. First, edge features represent only a small fraction of the image; using an extra path for edge supervision is still likely to provide insufficient information while generating redundant non-structural features \cite{MINet}. Second, the extra flows result in extra computational cost, making these methods less efficient.

The problem of low spatial coherence due to insufficient attention to inter-pixel relationships has been addressed by using post-processing methods such as conditional random fields (CRF) to refine the output prediction \cite{DSS,picanet,crf2}. However, these methods usually result in low processing speed. Some studies \cite{egnet,non-local} proposed that spatial coherence can be enhanced by adding complementary edge information. Other studies \cite{MINet,f3net} have suggested that the incoherence is caused by scale variation of multi-level features, and have proposed using multi-scale fusion to alleviate the problem.

Another approach to solve these problems is to remodel SOD with new informative labels. Traditional salient masks used as training labels treat all pixels inside a salient object equally and independently; as a result, they lack information about inter-pixel relationships and fundamentally make edges hard to detect. Therefore, using saliency masks as the only training label is a suboptimal choice. In the label decoupling framework (LDF) for SOD \cite{LDF}, traditional salient masks were decoupled into a location-aware detailed map and body map, which were used as auxiliary labels for training. However, these new labels required specifically designed extra supervision flows and were not proved to be compatible with other existing models. Another group \cite{ConnNet} introduced the connectivity mask—a multi-channel mask exhibiting connectivity of each pixel with its neighboring pixels—as the CNN label. Although the connectivity mask is an inter-pixel relation-aware label, this method completely replaces saliency prediction with pixel connectivity modeling, and therefore does not effectively utilize the original saliency information. In addition, the method ignores the inherent properties of this new label, making the results less promising. We propose that the design and effective utilization of an informative label which is \textit{compatible with any existing method} can efficiently improve the performance of existing models.

Inspired by this concept, we developed a novel connectivity-based SOD framework called the Bilateral Connectivity Network (BiconNet) as shown in Fig. \ref{overvew}. BiconNet consists of four parts: a connectivity-based SOD backbone, a bilateral voting (BV) module, a region-guided channel aggregation (RCA) module, and a bilateral connectivity (Bicon) loss function. To model inter-pixel relationships, we first replace the backbone’s label with a connectivity mask. Then, to enhance the spatial coherence between neighboring pixels, we use a BV module to obtain a more representative connectivity map called the Bicon map. After this step, we generate two single-channel saliency maps, with edge information emphasized, via an RCA module. Finally, we propose the Bicon loss function to further emphasize edge features and spatial consistency for final salient object detection.

\textbf{BiconNet exhibits three advantages}: First, by changing the CNN’s intermediate goal to predicting pixel-wise connectivity, inter-pixel relation modeling has become one of the network’s tasks. Thus, BiconNet can focus more attention on inter-pixel relationships. Second, based on the inherent property of connectivity masks, edge regions can be located directly from ground truth, which are then emphasized in the final output for network training via the RCA module. Compared to other edge-based methods \cite{egnet,poolnet}, this is a more efficient way to aggregate edge features. Third and most importantly, since BiconNet changes only the output layer of the backbones and all other modules (BV and RCA) are trained after it, BiconNet can be built on any saliency-based SOD framework without changing the framework’s original design (e.g., internal structure and loss functions), and will improve its performance. 

In summary, there are three main contributions of this work:

\begin{itemize}
\item We propose a connectivity-based SOD framework called BiconNet to explicitly model pixel connectivity, enhance edge modeling, and preserve spatial coherence of salient regions. BiconNet can be \textit{easily plugged into any existing SOD model} with neglectable parameter increases. 

\item We propose an efficient, connectivity-based edge feature extraction method that can directly emphasize the edge-specific information from the network output. We also introduce a new loss function, Bicon loss, to further enhance the utilization of the edge features and preserve the spatial consistency of the output.  

\item We build BiconNets with backbones of seven state-of-the-art SOD models. By comparing these BiconNets with the corresponding baselines, we show that our model outperforms the latter models on five widely used benchmarks using different evaluation metrics. 

\end{itemize}

\section{Related Work}
Earlier SOD methods \cite{tradition1,tradition_PR2,tradition2} mostly utilized hand-crafted features to detect salient regions. These methods cannot effectively capture high-level semantic information from data, and are ineffective when dealing with complex scenes in images. CNN-based models have recently become the main choice for SOD due to their multi-level feature extraction ability. However, in earlier CNN-based SOD models \cite{DSS,old2}, erroneous predictions were usually made near the salient edges, and low spatial coherence occurred in the middle of the salient region or near the edges. There are three ways to solve these problems: multi-scale feature aggregation models, edge-enhanced models, and problem remodeling methods.
\subsection{Multi-scale Feature Aggregation Models}

One reason for the problems described above is that detailed features can be diluted as the CNN becomes deeper. To utilize saliency features more efficiently, one solution is to aggregate multi-scale information. 
Hou et al. \cite{DSS} demonstrated that using short connections between different layers helped aggregate multi-scale features. Chen et al. \cite{gcpa} proposed a model that can aggregate low-level detailed features, high-level semantic features, and global context features to learn the relationship between different salient regions. Qin et al. \cite{U2-Net} proposed a nested network that uses Residual U-blocks to extracted multi-scale features. Li et al. \cite{HKU} extracted saliency features from three different scales of the images and aggregated them for final detection. Pang et al. \cite{MINet} extracted effective multi-scale features from two interaction modules and preserved the spatial consistency of intra-class units. Although effective, these methods usually require extra computational power for the frequent feature aggregations between different layers. 

\begin{figure}[h!]
\begin{center}
\includegraphics[width=1\linewidth]{{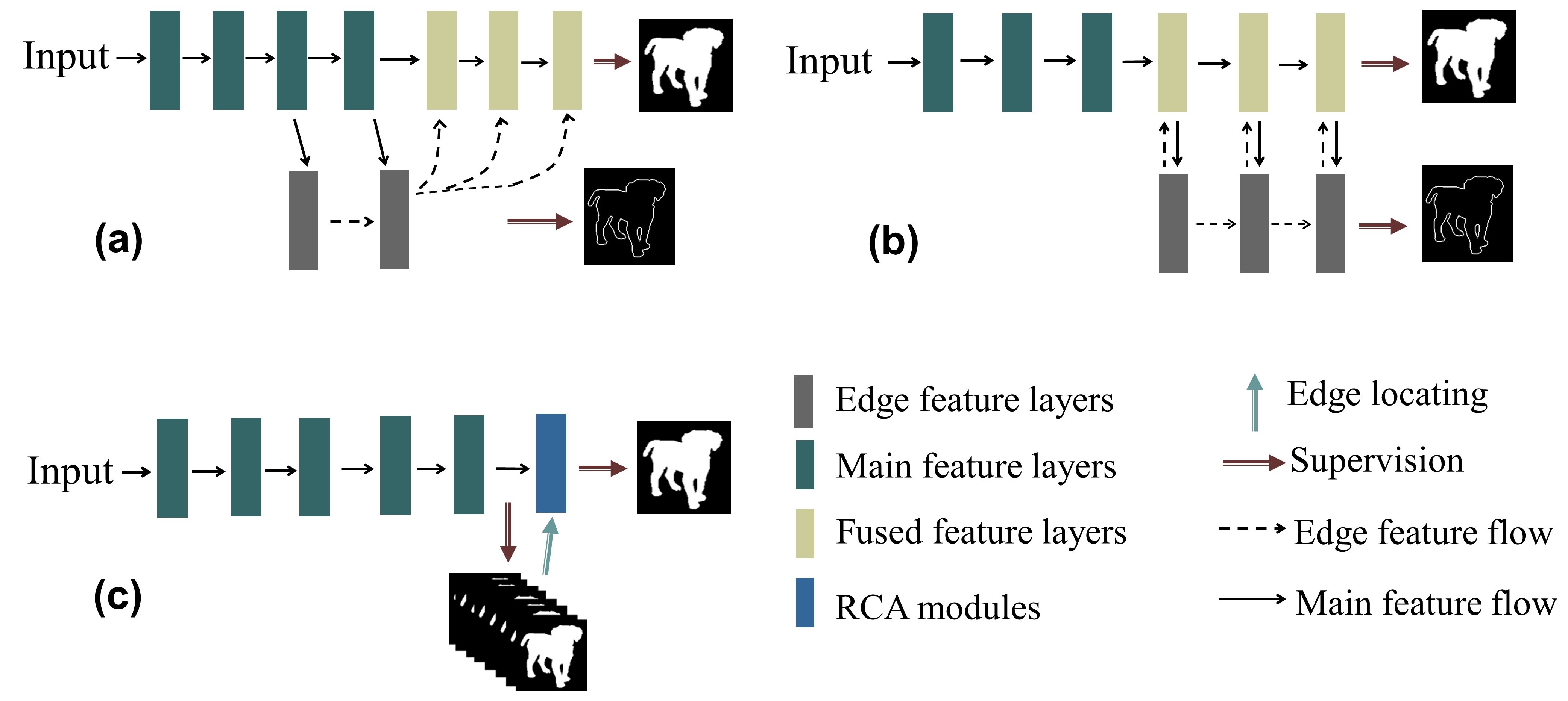}}
\end{center}
\vspace{-8pt}
   \caption{Different edge-based models: (a) edge cue models \cite{egnet,poolnet}; (b) interactive edge models \cite{scrn,ITSD}; (c) BiconNet. Both (a) and (b) need to include at least one extra flow branch for the edge features. In contrast, BiconNet can directly receive the edge location from the connectivity ground truth and then emphasize the edge-specific information in the output via a simple RCA module. }
\vspace{-20pt}
\label{edge_models}
\end{figure}

\subsection{Edge-enhanced Models}
To preserve edge information, edge-enhanced models intentionally generate extra edge features for training. Zhao et al. \cite{egnet} built another supervision flow for the edge features, which were fused with the salient features at the final stages of their network. Liu et al. \cite{poolnet} extracted the edge features from another edge detection dataset and used these for joint training with saliency detection. Qin et al. \cite{basnet} added a refinement module after their encoder-decoder structure to refine the boundary details. Zhang et al. \cite{bound_PR} proposed a boundary localization module to extract structural information.
Wu et al. \cite{scrn} exploited the logical interrelation between the edge map and saliency map and proposed a bidirectional framework to refine both tasks. 
Zhou et al. \cite{ITSD} proposed an approach that interactively fuse edge features and saliency features. These models show the effectiveness of adding edge features for saliency detection, but they usually generate redundant features and are computationally expensive since they add extra supervision flows for the edge path. In our work, the edge information is used in a more efficient way, as shown in Fig. \ref{edge_models}.

\subsection{Problem Remodeling Methods}
Compared to the above models which focus on the internal structure of the network, an efficient way to solve the SOD problem is to rethink the task and remodel it using more informative labels. Wei et al. \cite{LDF} decoupled the ground truth label into a body map and a detail map according to the location of object edges and used three supervision flows for training. However, the authors did not demonstrate a general way to utilize these labels in an existing framework. In addition, although these labels worked well in detecting the salient edges, they were not inter-pixel relation-aware. Kampffmeyer et al. \cite{ConnNet} replaced the saliency labels with connectivity masks and illustrated improvements achieved by this change. This approach, called ConnNet, remodeled the problem of SOD by converting the saliency prediction task into several sub-tasks of foreground connectivity prediction. However, this method did not fully utilize the information of the connectivity mask. In addition, the method is incompatible with many saliency evaluation metrics as it does not predict a single-channel saliency probability map. We propose a method to overcome these problems, described in the next sections.

\section{Proposed Method}
\subsection{Framework Overview}
Our framework, BiconNet, consists of four parts: a connectivity-based SOD backbone, a BV module, an RCA module, and a Bicon loss function. For the backbone, we can use any existing saliency-based SOD framework. An overview of our method is shown in Fig. \ref{overvew}.

\subsection{Connectivity Vector/Mask}
Given an existing SOD backbone, our first step is to replace its single-channel saliency map output with an 8-channel connectivity map by changing its fully connected layers and to replace its label with the connectivity mask. In the next step, we will introduce connectivity vectors and masks/maps. 

\begin{figure}[h!]
\begin{center}
   \includegraphics[width=1\linewidth]{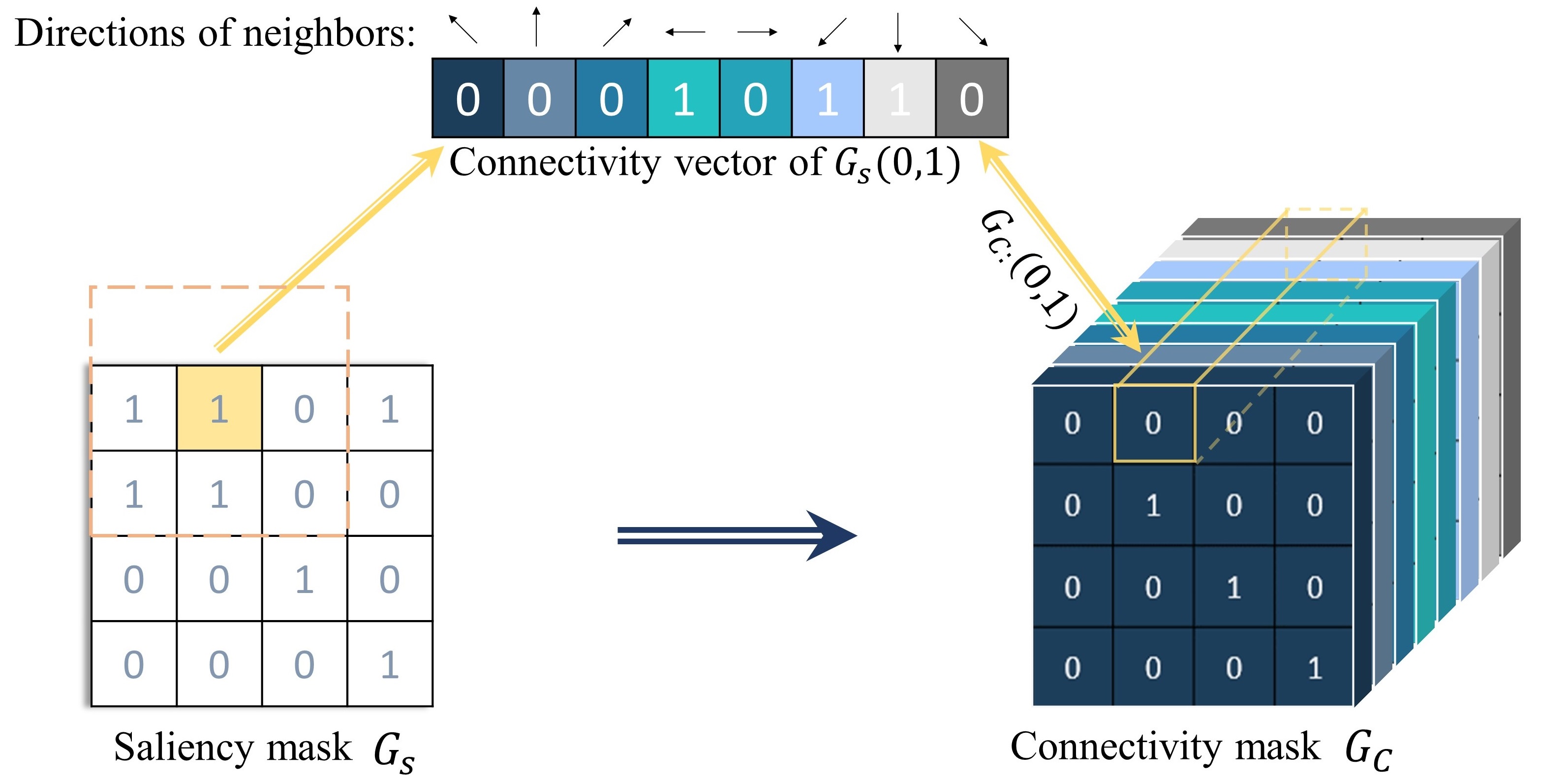}
\end{center}
\vspace{-8pt}
   \caption{Visualization of converting a saliency mask to a connectivity mask. The dashed box on $G_S$ shows the 8-neighbor region of the selected pixel. $G_S$ will be boundary-mirrored if needed. Every channel of $G_C$ represents pixel connectivity at a certain direction.}
   \vspace{-10pt}
\label{conversion}
\end{figure}

A connectivity \cite{connectivity} vector of a pixel is a multi-entry binary vector used to indicate whether the pixel is connected to its neighboring pixels. In the 8-neighbor system, given a pixel at coordinates  $(x,y)$, we use an 8-entry connectivity vector to represent the unidirectional connectivity with its neighbors in the square area of $[x\pm 1,y\pm 1]$ with every entry representing one specific direction. Given a binary saliency mask $G_S$ with size $H\times W$, by deriving the connectivity vector for every pixel in $G_S$, we obtain an 8-channel mask $G_C$ with size $H\times W \times 8$ called the connectivity mask (Fig. \ref{conversion}). The $i_{th}$ channel of $G_{C}$ ($G_{Ci}$) represents if the original pixels on $G_S$ are connected with their neighboring pixels at the $i_{th}$ directions (e.g., upper left if $i=1$ using row-major order). In this work, as in \cite{ConnNet}, We define connectedness only for the adjacent salient pixels. For better understanding, we call the discrete ground truths as connectivity masks $G_C$  and the network’s continuous outputs $C$ as connectivity maps.

\begin{figure}[h!]
\begin{center}
\includegraphics[width=0.75\linewidth]{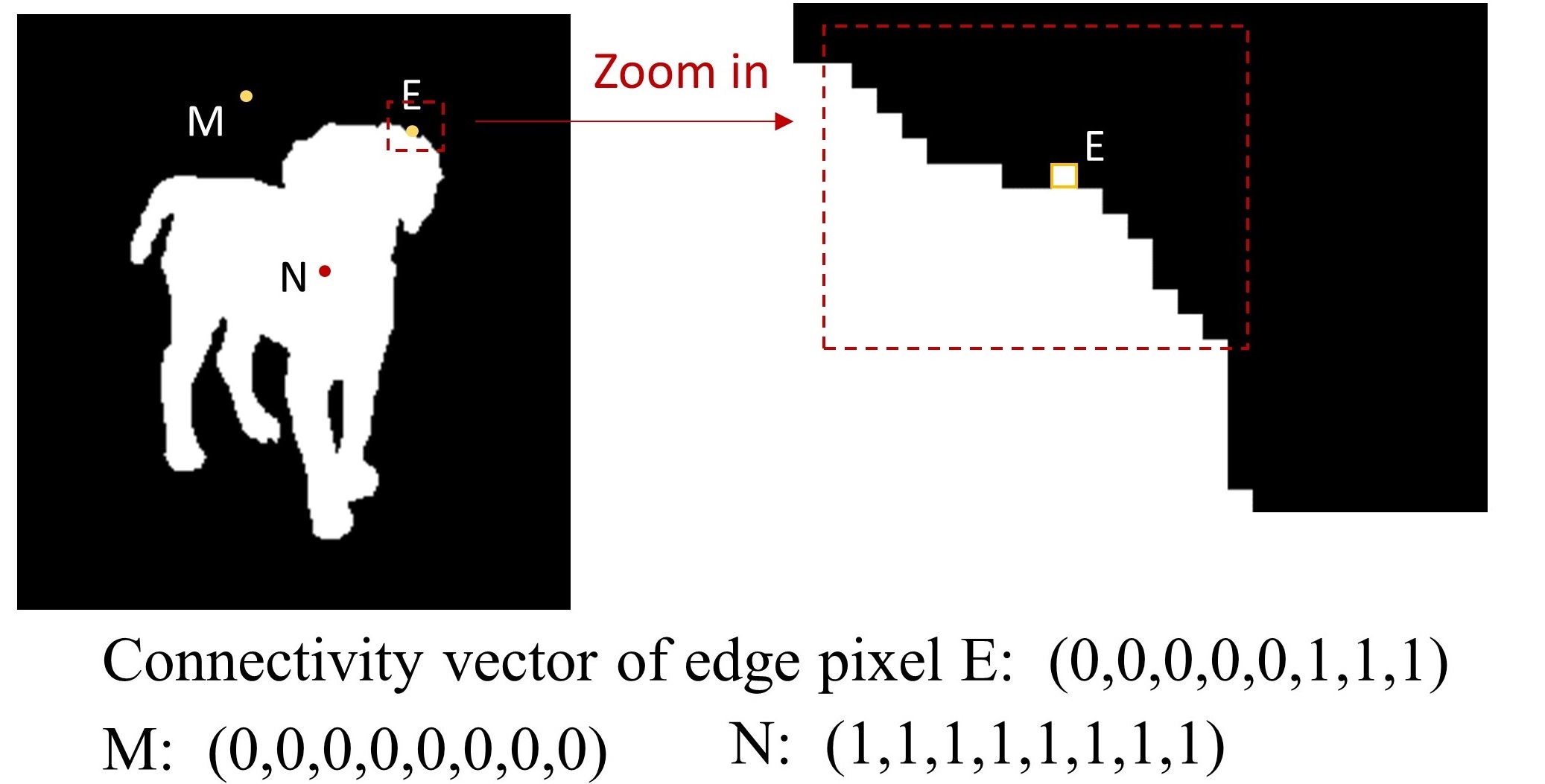}
\end{center}
\vspace{-8pt}
   \caption{The turbidity property for edge pixels. M, N, and E represent pixels in the background (M), inside the salient region (N), and at the edge (E). Only pixel E has a mixture of zeros and ones in its connectivity vector, whereas M and N have all-zeros and all-ones connectivity vectors, respectively.}
   \vspace{-10pt}
\label{turbidity}
\end{figure}

We show that learning a connectivity mask $G_C$ provides three main advantages over a binary segmentation mask $G_S$. First, compared to $G_S$ where every entry only indicates the saliency of the current pixel, $G_C$ focuses more on the mutual relationship between its pixels. Second, $G_C$ itself contains more structural information (such as edges) than $G_S$. Specifically, in $G_C$, the elements of the connectivity vector for an edge pixel are always a mixture of zeros and ones, whereas internal foreground pixels have all-ones connectivity vectors and background pixels have all-zeros connectivity vectors (Fig. \ref{turbidity}). We call this property the turbidity of the edge connectivity vectors. Thus, given a ground truth connectivity vector of a pixel, we can always determine whether it is an edge pixel 
simply by checking the zero and one distribution of the vector. As shown in future sections, this property is important as it provides an efficient way to utilize edge information. 
Third, besides showing the connectivity of saliency pixels, every entry of $G_C$ also reflects the connection direction. Thus, $G_C$ is a structure- and inter-pixel relationship-aware label.

\subsection{Bilateral Voting Module}

\begin{figure}[h!]
\begin{center}
   \includegraphics[width=1\linewidth]{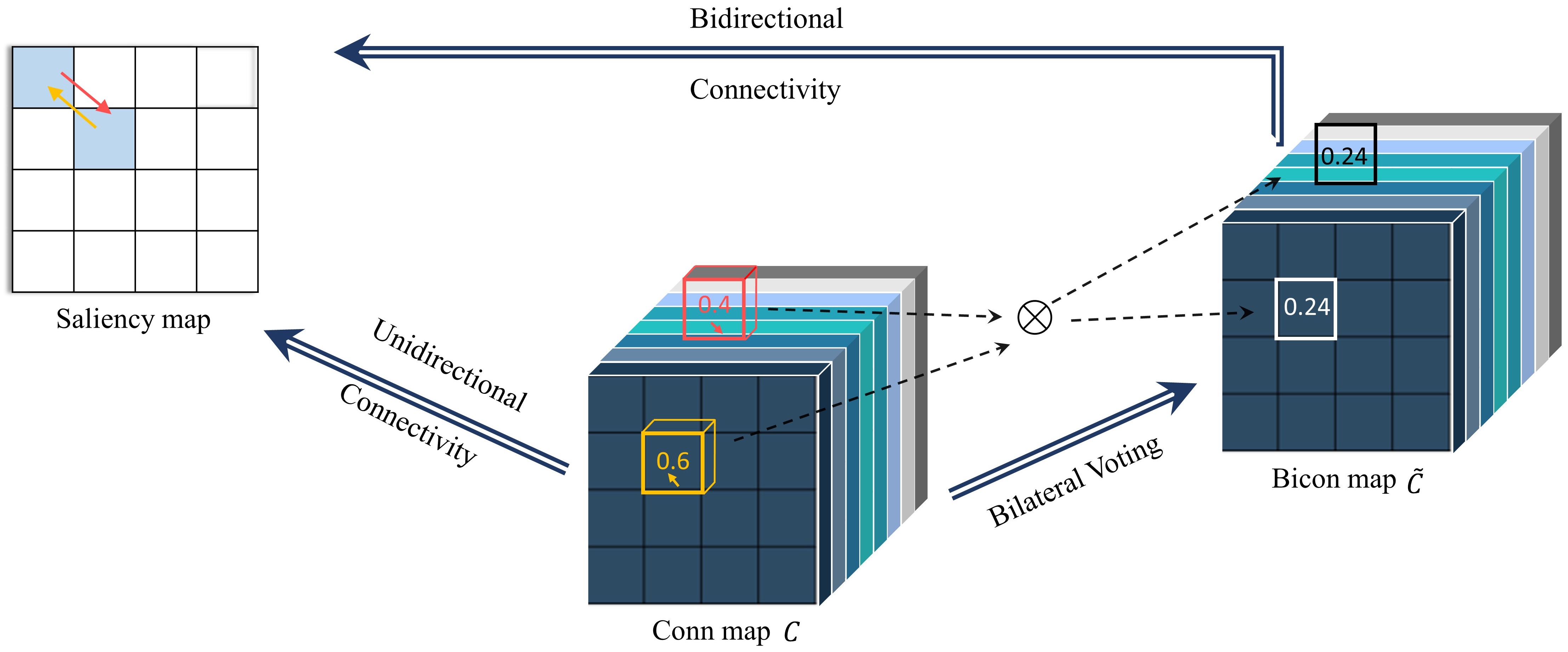}
\end{center}
\vspace{-20pt}
\caption{Illustration of how connectivity pair is defined and how a Bicon map is generated via bilateral voting (BV) when $a = 1$ and $b = 1$. In the predicted Conn map (middle) the two entries $C_1(2,2)$ and $C_8(1,1)$ represent two predicted unidirectional connectivity probabilities of the highlighted neighboring pixels at $(1,1)$ and $(2,2)$ in the saliency map (left). After the BV module, the generated Bicon map is the bidirectional representation of the connectivity for pixels in the saliency map. }
\label{bv}
\vspace{-10pt}
\end{figure}

For every two neighboring pixels in $G_S$, there is a unique element pair in $G_C$ representing the unidirectional connectivity between them. We call such a pair in $G_C$ a connectivity pair. To be specific, consider a pair of neighboring pixels on $G_S$: M at location $(x,y)$ and N at $(x+a,y+b)$ $a,b \in \{ 0,\pm 1\}$. We can obtain the unidirectional connectivity from M to N from the value of $G_{Cj} (x,y)$, where  $G_{Cj}$ is the channel that represents the relative direction from N to M. For example, if N is located lower-right of M, then $j = 8$ (row-major order). Similarly, the connectivity from N to M can be found at $G_{C(9-j)}(x+a,y+b)=G_{C1}(x+1,y+1)$. We call the two elements $G_{Cj} (x,y)$ and $G_{C(9-j)}(x+a,y+b)$ a connectivity pair of M and N. The same concept is also defined for the output connectivity map $C$, where every two neighboring pixels in the salient map have a unique connectivity pair in C representing the  \textit{probability} of the unidirectional connection. Fig. \ref{bv} shows an example of this case when $a=b=1$.

The concepts of saliency and connectedness are closely related and mutually convertible: If two pixels are connected, they are salient. Two pixels of $G_S$ are considered as connected (salient) if and only if both elements of its connectivity pair agree with this connection, i.e., if and only if $G_{Cj}(x,y)=G_{C(9-j)}(x+a,y+b)=1$.
We call this the \textit{discrete} bilateral connectivity agreement, which reveals the bidirectional property of pixel connections and shows the importance of mutual impacts between neighboring pixels.

From this agreement, we know theoretically that the two elements from a connectivity pair should have the same connection probability to each other. However, in practice, connectivity pairs of the network’s \textit{continuous} outputs (i.e., the connectivity maps $C$) rarely satisfy this agreement. These disagreements result in spatial inconsistencies. To model the neighboring dependency and preserve the spatial consistency, we propose a novel connectivity-enhancement module called bilateral voting (BV) module.

Given a connectivity map output $C$, the goal of the BV module is to generate another connectivity map that satisfies the bilateral connectivity agreement. To do so, we first extract all of the connectivity pairs. Then, we multiply the two elements in every connectivity pair and assign the resulting value to both elements, yielding a new connectivity map $\widetilde {C}$. This process is shown in Fig. \ref{bv} and is formulated as: 		
\begin{equation}
\begin{split}
{\widetilde C_j}(x,y) &= {\widetilde C_{9 - j}}(x + a,y + b) \\
&= {C_j}(x,y) \times {C_{9 - j}}(x + a,y + b),
\end{split}
\end{equation}
where the subscript j means the $j_{th}$ channel, $a,b \in \{ 0, \pm 1\}$. The logic behind the formula is that we can get the bidirectional pixel connection probability by multiplying every two elements of a connectivity pair, as each represents a unidirectional connectivity probability relative to the other. Since this continuous process is similar to making the discrete bilateral agreement, we call it bilateral voting. We name the new map $\widetilde {C}$ the bilateral connectivity map (Bicon map), and the original output $C$ the Conn map. In the later sections, we will show that the BV module is important both at the training and inference phases.

\subsection{Region-guided Channel Aggregation}
\label{section3.4}
After BV module, we obtain an enhanced 8-channel map $\widetilde {C}$ with every channel representing the bidirectional probability of pixel connection at a specific direction. In the previous sections, we pointed out that pixel connectivity is the sufficient and necessary condition of pixel saliency for neighboring pixels. Therefore, the overall probability of a pixel being connected with its neighbors reflects its saliency. To obtain a single-channel map representing saliency, we propose a region-guided channel aggregation (RCA) module to summarize the directional connectivity information in the eight channels of $\widetilde {C}$ into a single-channel output $\widetilde {S}$ using a function $f$. The generic form is written as:
\begin{equation}
\widetilde S(x,y) = f\{ {\widetilde C_i}(x,y)\} _{i = 1}^8\ \ ,
\end{equation}
where $f$ is an adaptive aggregating operation that varies with location $(x,y)$, $\widetilde {S}$ represents the aggregated overall probability of current pixel being salient. This process can also be interpreted as applying a function $f$ over every predicted connectivity vector in $\widetilde {C}$ to obtain an overall connection probability for the corresponding pixel. Here we define two types of $f$:

\textbf{Global aggregation.} A simple way to aggregate the values from different directions into a single value is to average them. By doing this, we will obtain a single-channel map with every pixel representing the average connection probability to its neighbors. In this case, $f$ is the averaging operation for all locations. We call the resultant map the global map, denoted as $\widetilde S_{global}$:
\begin{equation}
{\widetilde S_{global}}(x,y) = Mean\{ {\widetilde C_i}(x,y)\} _{i = 1}^8\ \ .
\end{equation}

\textbf{Edge-guided aggregation.} As mentioned, the edge pixels are the only pixels that have zero-one ground truth connectivity vectors. This property yields two advantages. First, we can generate ground truth edge masks by simply searching the location of zero-one vectors in connectivity masks. Based on these ground truth edge masks, we can locate and decouple the edge regions and non-edge regions from the output. Second, we can encourage the network to learn this special data representation based on prior knowledge about the turbidity of edge vectors. Due to the imbalance between edge pixels and non-edge pixels, the network intends to make uniform predictions among all directions; i.e., predicting connectivity vectors as all ones or all zeros. An edge pixel, since it is part of the salient region, is more likely to be predicted as an internal foreground pixel with an all-ones connectivity vector. This is the main reason for blurring edges: it is difficult for networks to learn to discriminate edge pixels from other salient pixels. To emphasize the difference between these two types of pixels, we want the networks to pay extra attention to the likely misclassified direction of a predicted edge connectivity vector; i.e., directions that are in fact not connected to the current edge pixel. As for non-edge pixels, since they have all-ones or all-zeros ground truth connectivity vectors, we want the network to uniformly focus on all directions. To this end, we designed a region-adaptive aggregation method for these two regions:
\begin{equation}
\label{decouple}
\Scale[0.85]{
\widetilde S_{decouple}(x,y)=
\begin{cases}
{1 - \min \{ {{\widetilde C}_i}(x,y)\} _{i = 1}^8}& {(x,y) \in {P_{edge}}},\\
{Mean\{ {{\widetilde C}_i}(x,y)\} _{i = 1}^8}& {(x,y) \notin {P_{edge}}},
\end{cases}}
\end{equation}
where $\widetilde S_{decouple}$ is called the edge-decoupled map and $P_{edge}$ is the set of edge pixels. For the edge part, we emphasized the most likely disconnected directions by finding the minimum values of the predicted connectivity vectors. This design is highly correlated with the loss function, which is discussed in the next section.

So far, we have generated two single-channel maps: the global map $\widetilde S_{global}$ and the edge-decoupled map $\widetilde S_{decouple}$ as shown in Fig. \ref{overvew}. $\widetilde S_{decouple}$ is used for learning the edge-specific information; $\widetilde S_{global}$ is a more robust representation of salient objects and will be used as the final saliency prediction during inference.

\subsection{Bicon Loss}
Our loss function is defined as:
\begin{equation}
{L_{bicon}} = {L_{decouple}} + {L_{con\_const}} + {L_{opt}}\ \ .
\end{equation}

We call this hybrid loss the Bicon loss $L_{bicon}$, where $L_{decouple}$ is the edge-decoupled loss, $L_{con\_const}$ is the connectivity consistency loss, and $L_{opt}$ is the optional loss. We define the loss terms in the following sections.

\textbf{Edge-decoupled loss.} Binary cross entropy (BCE) \cite{BCE} is one of the most widely used losses function, defined as:
\begin{equation}
\vspace{-3pt}
\Scale[0.66]{ L_{bce}(S,G) = -\sum\limits_{(x,y)} {[G(x,y)*\log (} {S}(x,y)) + (1 - {G}(x,y))*\log (1 - S(x,y))],}
\end{equation}
where $G(x,y)\in \{0,1\}$ is the ground truth label of pixel $(x,y)$ and $S(x,y)$ is the prediction. BCE loss is a pixel-wise loss function that considers each pixel equally, thus it does not consider inter-pixel relationships when the traditional saliency maps are used as the loss input \cite{MINet,f-loss,basnet}. We propose that this problem can be alleviated with use of a more informative and spatial relation-aware input. To this end, we used $\widetilde S_{decouple}$ as the input of BCE. Although BCE is still calculating the loss independently for every unit, single units carry information about their intrinsic saliency scores and the region-based connectivity. Based on Eq. \ref{decouple}, this loss is formulated as:

\begin{equation}
\Scale[0.85]{
\begin{aligned}
&{L_{decouple}} = {L_{bce}}({\widetilde S_{decouple}},{G_S})\\
& =\begin{cases}
{{L_{bce}}(1 - \min \{ {{\widetilde C}_i}(x,y)\} _{i = 1}^8,{G_S}(x,y))} &{(x,y)\in {P_{edge}}},\\
{{L_{bce}}(mean\{ {{\widetilde C}_i}(x,y)\} _{i = 1}^8,{G_S}(x,y))}& {(x,y) \notin {P_{edge}}},
\end{cases}
\end{aligned}}
\end{equation}

where  $G_S(x,y)\in \{0,1\}$ is the saliency ground truth label of pixel $(x,y)$, indicating whether the pixel is salient. Specifically, we can derive the edge part as:
\begin{equation}
\Scale[0.95]{
\begin{array}{l}
\;\;\;{L_{bce}}(1 - \min \{ {\widetilde C_i}(x,y)\} _{i = 1}^8,{G_S}(x,y))\\
 = {L_{bce}}(1 - \min \{ {\widetilde C_i}(x,y)\} _{i = 1}^8,\;{\textbf{1}})\\ 
 = {L_{bce}}(\min \{ {\widetilde C_i}(x,y)\} _{i = 1}^8,\;{\textbf{0}})\ \ .
\end{array}}
\end{equation}

For the edge pixels, our goal is to make the network learn the sparse representation of the turbid edge vectors. As previously discussed, the edge pixels are most likely to be predicted as internal salient pixels that have all ones in their connectivity vectors. Thus, a feasible way to learn the turbidity is to force the minimum value of the edge connectivity vector to be zero; i.e., we want the network to only focus on the disconnected direction for edge pixels. For the non-edge pixels, since they all have all-zeros or all-ones connectivity vectors, our goal is to make the average value across channels to be close to their labels. Namely, we want the network to put uniform weights among all directions.

\textbf{Connectivity consistency loss.} The connectivity consistency loss is the weighted sum of BCE losses applied to both the original Conn map ($C$) and the Bicon map ($\widetilde{C}$). It is defined as:
\begin{equation}
\Scale[0.95]{
\begin{split}
{L_{con\_const}} &= {\omega _1}*{L_{conmap}} + {\omega _2}*{L_{bimap}}\\
 &= {\omega _1}*{L_{bce}}(C,\;{G_C}) + {\omega _2}*{L_{bce}}(\widetilde C,\;{G_C}),
\end{split}}
\end{equation}
where $G_C$ is the corresponding ground truth 8-channel connectivity map with every element $G_{Ci}(x,y) \in \{ 0,1\}$, specifying whether a pixel at location $(x,y)$ is connected to its specific neighboring pixel. $\omega _1$ and $\omega _2$ are weighting factors. The first term, $L_{conmap}$, is designed for preserving spatial consistency. For the second term, since the bidirectional connection probability in $\widetilde{C}$ is exponentially correlated with the original unidirectional probability, it usually generates larger loss on the hard pixels \cite{f3net}, such as edge pixels, while generating a smaller one on the easy pixels. Thus, it puts more weights on the object edges and helps maintain structural consistency as in \cite{f3net}. Furthermore, we consider $\widetilde{C}$ as the set of the equalized connectivity pairs so that intuitively $L_{bimap}$ is a ‘pair-wise’ loss which computes the loss of every pair in $\widetilde{C}$ twice. Thus, it should have a lower weight. For all of our experiments, we set $\omega _1=0.8$, $\omega _2=0.2$ unless otherwise noted.

\textbf{Optional loss.} As mentioned above, the BV and RCA modules together with the Bicon loss can be inserted into any existing saliency-based backbone to form the BiconNet architecture. Some existing studies \cite{basnet,MINet} have proposed specific loss functions with their network architectures. To maintain the integrity of these backbones, we apply the same loss function in these papers as our third term:
\begin{equation}
{L_{opt}} = {L_{orig}}({\widetilde S_{global}},{G_S}),
\end{equation}
where $L_{orig}(\cdot)$ is the loss function defined in the original backbone’s paper, $\widetilde{S}_{global}$ is the global map. Note that $L_{opt}$ is an optional loss term and will be applied according to the selection of backbones. 
\subsection{Inference}
To obtain the single-channel saliency probability map in the inference stage of BiconNet, we first pass the output Conn map $C$ through the BV module to get the Bicon map $\widetilde{C}$. Then, we aggregate the channels with the averaging operation to get the global map $\widetilde S_{global}$. Finally, we use $\widetilde S_{global}$ as the predicted saliency map, as shown in Fig. \ref{overvew}.

\section{Experiments}
\subsection{Datasets and Evaluation Metrics}
We evaluated our model on five frequently used SOD benchmark datasets: HKU-IS \cite{HKU} with 4,447 images, DUTS \cite{duts} with 10,553 images for training (DUTS-TR) and 5,019 for testing (DUTS-TE), ECSSD \cite{ecssd} with 1,000 images, PASCAL-S \cite{pascal} with 850 images, and DUT-OMRON \cite{omron} with 5,168 images. For the evaluation metrics, we adopted the mean absolute error (MAE) \cite{mae}, F-measure ($F_\beta$)  \cite{F-measure}, and E-measure ($E_m$) \cite{Emeasure}.  For the F-measure, we used the mean F-measure, $F_{ave}$, which is generated by thresholding the prediction map using an adaptive value equal to twice the mean of the prediction and is correlated with spatial consistency of the prediction \cite{cpd}.


\subsection{Experiment Setup and Implementation Details}
\textbf{Model Setup.} We adopted seven state-of-the-art models as both baselines and backbones to form the BiconNets: PoolNet \cite{poolnet}, CPD-R \cite{cpd}, EGNet \cite{egnet}, F3Net \cite{f3net} ,GCPANet \cite{gcpa}, ITSD \cite{ITSD}, MINet \cite{MINet}. We replaced all of their saliency prediction layers with 8-channel fully-connected layers, followed by our BV and RCA modules. We used Bicon Loss as the loss function for all models. For the models with deep supervision mechanisms such as \cite{egnet,gcpa}, we replaced all of the fully-connected layers with our connectivity layer followed by BV and RCA. For the extra edge supervision flows in \cite{egnet,ITSD}, we only replaced their edge labels with our connectivity-based edge labels generated by zero-one vector searching as discussed in Section \ref{section3.4} for consistency.

\textbf{Implementation Details.} We used the released official codes of the backbones for training both the baselines and the BiconNets. For baselines, we trained all of them from scratch, strictly following the instructions on their websites and the hyperparameter setting in their original papers. For the BiconNets, we used the same data pre-processing tricks as the corresponding baselines. For the hyperparameters, we only changed the starting learning rate (about 40\% of the baselines') and the batch size for our BiconNets, as in Table \ref{hyper}. The rest of hyperparameters were the same as the baselines’. We implemented all our experiments in Pytorch 1.4.0 \cite{pytorch} using an NVIDIA RTX 2080Ti GPU. The code is available at: https://github.com/Zyun-Y/BiconNets.

\begin{table*}[h!]
\renewcommand\arraystretch{1.2}
\centering
\setlength{\tabcolsep}{1mm}{
\small
\caption{The starting learning rate and batch size of BiconNet with different backbones.}
\label{hyper}
\begin{tabular}{cccccccc} 
\hline
Backbone&PoolNet&CPD-R&EGNet&F3Net&GCPANet&ITSD&MINet\\
\hline
Start Lr&$2e^{-4}$&$3.5e^{-5}$&$2e^{-5}$&$0.0018$&$0.01$&$0.005$&$0.0018$\\ 
Batch Size&$10$&$10$&$10$&$16$&$16$&$8$&$32$\\ 
\hline
\end{tabular}}
\end{table*}

\begin{table*}[h!]
\renewcommand\arraystretch{1.2}
\centering
\setlength{\tabcolsep}{0.2mm}{
\caption{Quantitative evaluation. seven methods were tested among five benchmark datasets. The mean F-measure ($F_{ave}$), mean absolute error (MAE), and E-measure ($E_m$) were used to evaluate the results. ↑ indicates that higher is better. We highlight the better result between every baseline and its BiconNet in {\color{BrickRed}red}. We denote the best result of a column with a {\color{blue}$\dagger$} superscript, the second best one with a {\color{blue}$\ast$} superscript.}
\label{results}
\resizebox{\textwidth}{43mm}{
\begin{tabular}{c ccc ccc ccc ccc ccc}
\hline
\multirow{2}*{Model}&\multicolumn{3}{c}{HKU-IS}&\multicolumn{3}{c}{DUT-TE}&\multicolumn{3}{c}{DUT-OMRON}&\multicolumn{3}{c}{PASCAL-S}&\multicolumn{3}{c}{ECSSD}\\  

\cmidrule(lr){2-4} \cmidrule(lr){5-7} \cmidrule(lr){8-10}\cmidrule(lr){11-13} \cmidrule(lr){14-16} 
&F$_{ave}\uparrow$&MAE$\downarrow$&$E_m\uparrow$&F$_{ave}\uparrow$&MAE$\downarrow$&$E_m\uparrow$&F$_{ave}\uparrow$&MAE$\downarrow$&$E_m \uparrow$&F$_{ave}\uparrow$&MAE$\downarrow$&$E_m\uparrow$&F$_{ave}\uparrow$&MAE $\downarrow$&$E_m\uparrow$\\

\hline

{PoolNet$_{19}$ \cite{poolnet}} &0.885&0.038&0.941&0.787&0.047&0.876&0.728&0.061&0.851&0.787&0.085&0.833&0.904&0.045&0.919\\ 
{PoolNet + Bicon}&\color{BrickRed}0.909&\color{BrickRed}0.034&\color{BrickRed}0.950&\color{BrickRed}0.826&\color{BrickRed}0.042&\color{BrickRed}0.902&\color{BrickRed}0.759&\color{BrickRed}0.057&\color{BrickRed}0.866&\color{BrickRed}0.812&\color{BrickRed}0.072&\color{BrickRed}0.853&\color{BrickRed}0.916&\color{BrickRed}0.040&\color{BrickRed}0.925\\ 
\cline{2-7}  

\cline{1-16}
{CPD-R$_{19}$ \cite{cpd}}&0.888&0.034&0.946&0.788&0.044&0.886&0.737&0.056&0.863&0.783&0.071&0.848&0.892&\color{BrickRed}0.038&0.925\\
{CPD-R + Bicon}&\color{BrickRed}0.905&0.034&\color{BrickRed}0.952&\color{BrickRed}0.806&0.044&\color{BrickRed}0.895&\color{BrickRed}0.750&0.056&\color{BrickRed}0.867&\color{BrickRed}0.794&\color{BrickRed}0.069&\color{BrickRed}0.857&\color{BrickRed}0.898&0.039&0.925\\

\cline{1-16}
{EGNet$_{19}$ \cite{egnet}}&0.900&0.031&0.952&0.804&0.038&0.894&0.750&0.053&0.867&0.794&0.073&0.847&0.905&0.037&0.927\\
{EGNet + Bicon}&\color{BrickRed}0.917&0.031&\color{BrickRed}0.954&\color{BrickRed}0.842\color{blue}$^\ast$&\color{BrickRed}0.037\color{blue}$^\ast$&\color{BrickRed}0.912\color{blue}$^\ast$&\color{BrickRed}0.770&\color{BrickRed}0.050\color{blue}$^\dagger$&\color{BrickRed}0.868&\color{BrickRed}0.821&\color{BrickRed}0.067&\color{BrickRed}0.863\color{blue}$^\ast$\color{BrickRed}&\color{BrickRed}0.922&0.037&\color{BrickRed}0.930\color{blue}$^\dagger$\\

\cline{1-16}
{F3Net$_{20}$ \cite{f3net}}&0.914&0.031&0.953&0.828&0.039&0.896&0.749&0.055&0.853&0.830&0.062&\color{BrickRed}0.857&0.924&0.037&0.926\\

{F3Net + Bicon}&\color{BrickRed}0.915&\color{BrickRed}0.029&\color{BrickRed}0.954&\color{BrickRed}0.835&\color{BrickRed}0.038&\color{BrickRed}0.899&\color{BrickRed}0.765&\color{BrickRed}0.051\color{blue}$^\ast$&\color{BrickRed}0.863&0.830&0.062\color{blue}$^\ast$&0.855&\color{BrickRed}0.927&\color{BrickRed}0.034\color{blue}$^\dagger$&\color{BrickRed}0.929\color{blue}$^\ast$\\

\cline{1-16}
{GCPANet$_{20}$ \cite{gcpa}}&0.896&0.032&0.950&0.812&\color{BrickRed}0.038&0.892&0.743&0.056&0.856&0.812&0.063\color{blue}$^\ast$&0.845&0.913&\color{BrickRed}0.035&0.924\\

{GCPANet + Bicon}&\color{BrickRed}0.918\color{blue}$^\ast$&0.032&\color{BrickRed}0.954&\color{BrickRed}0.834&0.040&\color{BrickRed}0.901&\color{BrickRed}0.762&\color{BrickRed}0.055&\color{BrickRed}0.863&\color{BrickRed}0.838\color{blue}$^\ast$&\color{BrickRed}0.061\color{blue}$^\dagger$&\color{BrickRed}0.858&\color{BrickRed}0.929\color{blue}$^\ast$&0.036&\color{BrickRed}0.929\color{blue}$^\ast$\\

\cline{1-16}
{ITSD$_{20}$ \cite{ITSD}}&0.900&0.030&0.952&0.806&0.041&0.891&0.752&0.058&0.862&0.800&0.067&0.850&0.903&\color{BrickRed}0.034\color{blue}$^\dagger$&0.925\\

{ITSD + Bicon}&\color{BrickRed}0.908&\color{BrickRed}0.029&0.952&\color{BrickRed}0.838&\color{BrickRed}0.038&\color{BrickRed}0.905&\color{BrickRed}0.774\color{blue}$^\ast$&\color{BrickRed}0.053&\color{BrickRed}0.874\color{blue}$^\ast$&\color{BrickRed}0.831&\color{BrickRed}0.064&\color{BrickRed}0.857&\color{BrickRed}0.920&0.035\color{blue}$^\ast$&\color{BrickRed}0.926\\
\cline{1-16}

{MINet$_{20}$ \cite{MINet}}&0.916&\color{BrickRed}0.026\color{blue}$^\dagger$&0.956\color{blue}$^\ast$&0.838&0.035\color{blue}$^\dagger$&0.903&0.762&0.053&0.870&0.830&0.064&0.858&0.926&\color{BrickRed}0.035\color{blue}$^\ast$&0.924\\

{MINet + Bicon}&\color{BrickRed}0.923\color{blue}$^\dagger$&0.028\color{blue}$^\ast$&\color{BrickRed}0.957\color{blue}$^\dagger$&\color{BrickRed}0.856\color{blue}$^\dagger$&0.035\color{blue}$^\dagger$&\color{BrickRed}0.915\color{blue}$^\dagger$&\color{BrickRed}0.778\color{blue}$^\dagger$&\color{BrickRed}0.051\color{blue}$^\ast$&\color{BrickRed}0.875\color{blue}$^\dagger$&\color{BrickRed}0.846\color{blue}$^\dagger$&\color{BrickRed}0.061\color{blue}$^\dagger$&\color{BrickRed}0.868\color{blue}$^\dagger$&\color{BrickRed}0.933\color{blue}$^\dagger$&0.036&\color{BrickRed}0.929\color{blue}$^\ast$\\
\hline
\end{tabular}}}
\vspace{-10pt}
\end{table*}

\subsection{Comparison with State-of-the-art Methods}
\textbf{Quantitative Comparison.} To compare our method and the baselines, we list all experiments and their results in Table \ref{results}. As the results show, the absolute majority of our results (98/105) show better or the same performance compared to the corresponding baselines. Our method also achieved most of the best overall results (14/15) (marked with {\color{blue}$\dagger$}). The results also indicate that our model can make a uniform prediction on the salient regions and preserve spatial consistency of the input more effectively than the baseline.

\begin{figure*}[h!]
\begin{center}
\includegraphics[width=1\linewidth]{{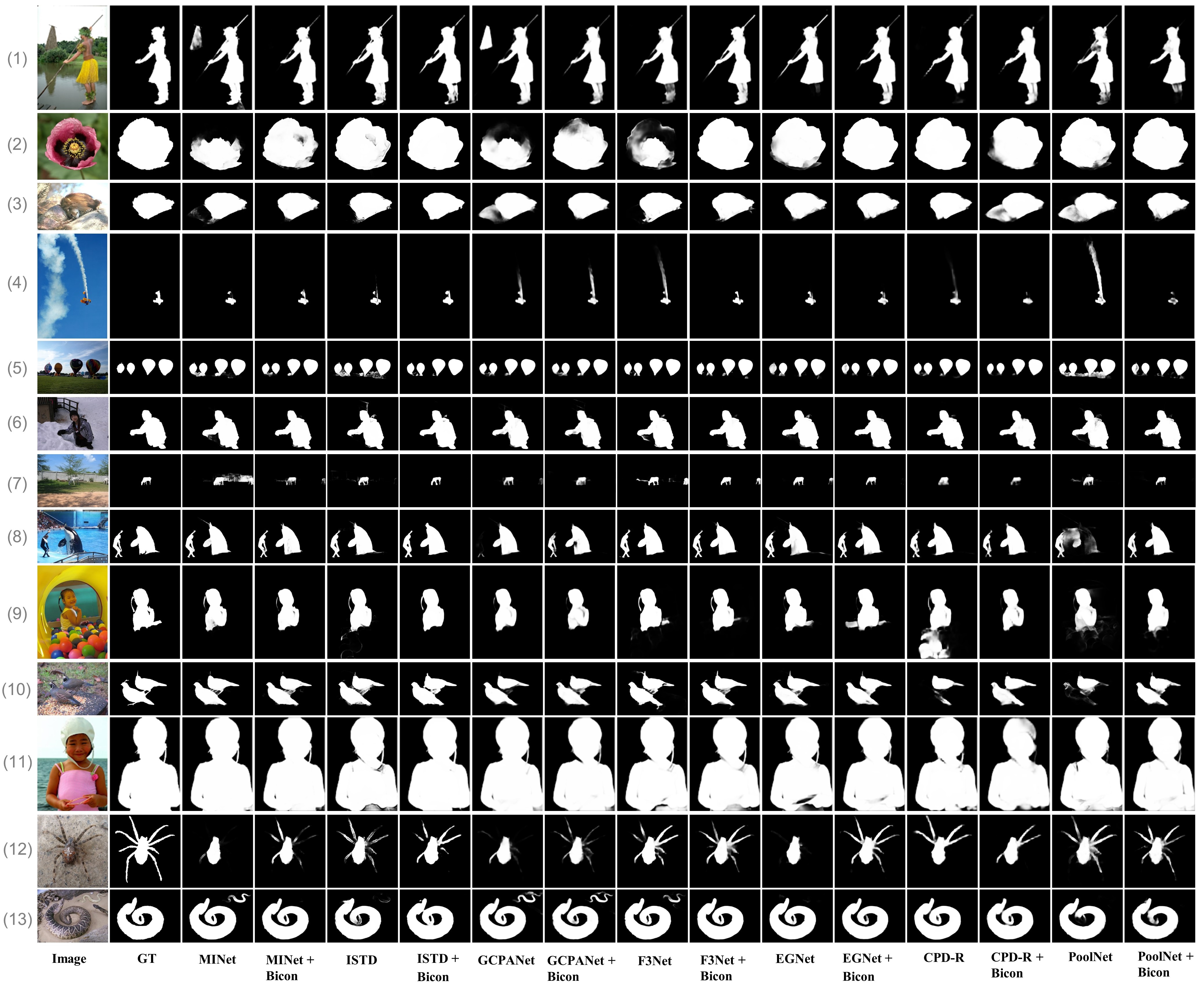}}
\end{center}
\vspace{-8pt}
   \caption{Visual comparisons of different models.}
\label{visualizations}
\end{figure*}

\textbf{Qualitative Evaluation.} Representative examples of our qualitative analyses are shown in Fig. \ref{visualizations}. Compared to baselines, our model can predict sharper boundaries and uniformly highlight salient regions in various challenging scenarios, including small objects (rows 4 and 7), complex background (rows 1, 3, 9 and 10) and foreground (rows 2 and 11), multiple objects (rows 5, 8 and 10), and interfering objects in the background (row 13).

\subsection{Ablation Study}
In this section, we study the effectiveness of different components of our model. The experiments in this section were trained on the DUT-TR dataset and tested on DUT-OMRON and HKU-IS. For a fair comparison, all experiments use GCPANet \cite{gcpa} as backbone. The overall ablation study results are listed in Table \ref{ablation}.

\begin{table*}[t]
\renewcommand\arraystretch{1.2}
\centering
\setlength{\tabcolsep}{1mm}{
\caption{Ablation study on HKU-IS and DUT-OMRON datasets.}
\label{ablation}
\begin{tabular}{cccccccccccccc} 
\hline
\multirow{2}*{Exp}&\multicolumn{7}{c}{Model}&\multicolumn{3}{c}{HKU-IS}&\multicolumn{3}{c}{DUT-OMRON}\\
\cmidrule(lr){2-8} \cmidrule(lr){9-11} \cmidrule(lr){12-14}
&$Base$&$Conn$&$L_{conmap}$&$BV$&$L_{global\_bce}$&\makecell[c]{$L_{decouple}$\\$(RCA)$}&$L_{con\_const}$
&F$_{ave}\uparrow$&MAE$\downarrow$&$E_m\uparrow$&F$_{ave}\uparrow$&MAE$\downarrow$&$E_m\uparrow$\\
\cmidrule(lr){1-8} \cmidrule(lr){9-11} \cmidrule(lr){12-14}
1&$\surd$&&&&&&&0.896&0.032&0.950&0.743&0.056&0.856\\

2& &$\surd$&$\surd$&&&&&0.899&0.033&0.949&0.738&0.058&0.854\\
3& &$\surd$&$\surd$&$\surd$&$\surd$&&&0.911&0.031&0.951&0.750&0.057&0.853\\
4& &$\surd$&$\surd$&$\surd$&&$\surd$&&0.916&0.033&0.951&0.760&0.057&0.860\\
5& &$\surd$& &$\surd$&&$\surd$&$\surd$&0.918&0.032&0.954&0.762&0.055&0.863\\

\hline

\end{tabular}}

\end{table*}

\textbf{Connectivity modeling.} We explore the role of the connectivity prediction strategy using two experiments. First, we used the original GCPANet as our baseline, denoted as $Base$ (Exp. 1). Then, we replaced its output layers with 8-channel connectivity prediction layers and used connectivity masks instead of the saliency masks as our ground truth. We denote this connectivity version of the baseline as $Conn$. For the loss function, we used the multi-channel BCE loss $L_{conmap}$ for the output Conn map $C$. This second experiment, denoted as Exp. 2 in Table \ref{ablation}, is very similar with ConnNet proposed in \cite{ConnNet}. We used channel averaging at testing to get the single-channel saliency maps for evaluation. As seen in Table \ref{ablation}, the results did not improve compared to Exp. 1, which follows our key hypothesis that completely replacing saliency modeling with connectivity modeling is not sufficient for modeling the saliency region.

\textbf{Bilateral voting mechanism.} Next, we studied the proposed BV module, which is important both at training and testing phases. The BV module helps the training in two ways: first, it provides an enhanced connectivity map $\widetilde{C}$ for the RCA module; second, in the connectivity consistency loss term, it generates the input for $L_{bimap}$, which is a position-aware loss. To simplify the experiment and avoid interference, we tested only the first part in this subsection. Based on $Conn$, we first conducted the bilateral voting on the output Conn map $C$ and got the Bicon map $\widetilde{C}$. Then, we computed the global map $\widetilde{S}_{global}$ by averaging among channels of $\widetilde{C}$. For the loss term, we calculated the BCE loss on both the global map ($L_{global\_bce}$) and the Conn map ($L_{conmap}$). This process is shown as Exp. 3 of Table \ref{ablation}. As seen, inclusion of the BV module improved the  $F_{ave}$, indicating that the BV module can enhance the spatial consistency of the output predictions.

\begin{figure}[h]
\begin{center}
   \includegraphics[width=1\linewidth]{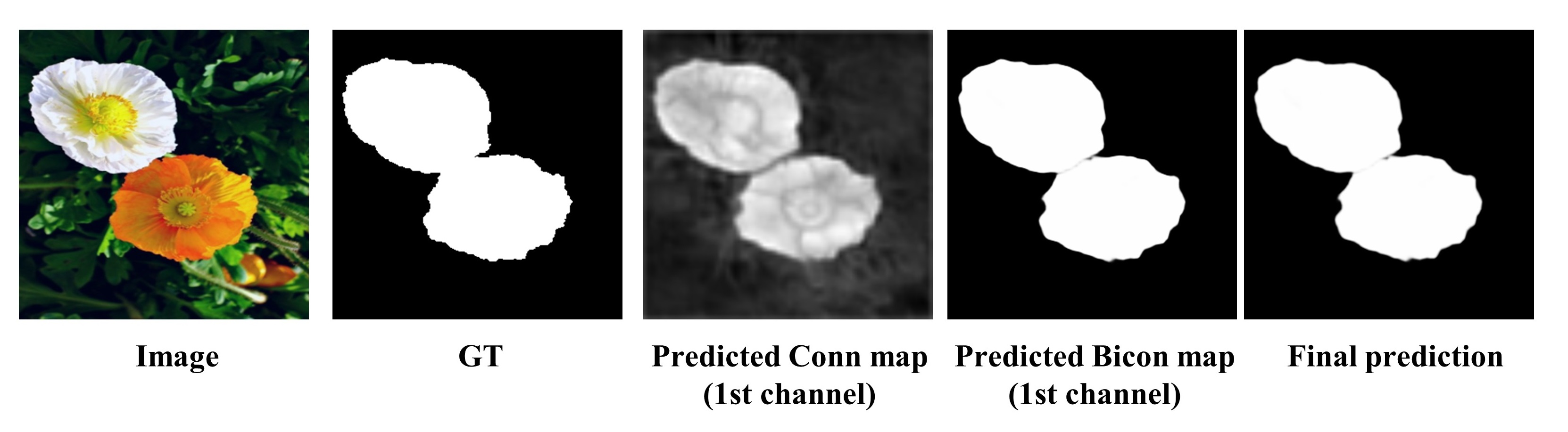}
\end{center}
\vspace{-8pt}
\caption{Visualization of outputs at different stages of BiconNet. As representative examples, for the Conn map $C$ and Bicon map $\widetilde{C}$, we show only the first channel. The predicted Bicon map has much higher spatial coherence than the Conn map.}
\label{BV_visualization}
\vspace{-10pt}
\end{figure}

To test the effectiveness of the BV module at the testing phase, based on Exp. 3, we tested the output both with and without the BV module. As seen in Table \ref{testing_method} and in Fig. \ref{BV_visualization}, all three metrics have been improved after we applied the BV module to the testing phase.

\begin{table}[h]
\renewcommand\arraystretch{1.2}
\centering
\setlength{\tabcolsep}{1.3mm}{
\small
\caption{Different testing methods based on Exp. 3.}
\label{testing_method}
\begin{tabular}{ccccccc} 
\hline
\multirow{2}*{Test Method}&\multicolumn{3}{c}{HKU-IS}&\multicolumn{3}{c}{DUT-OMRON}\\
\cmidrule(lr){2-4} \cmidrule(lr){5-7}
&F$_{ave}\uparrow$&MAE$\downarrow$&$E_m\uparrow$&F$_{ave}\uparrow$&MAE$\downarrow$&$E_m\uparrow$\\
\hline
Without BV&0.889&0.033&0.945&0.732&0.061&0.849\\ 
With BV&0.911&0.031&0.951&0.750&0.057&0.853\\ 
\hline

\end{tabular}}

\end{table}

\textbf{The edge decoupling mechanism.} In this subsection, we study the proposed edge decoupling mechanism, which is the key innovation of the RCA module. Based on Exp. 3, we computed the edge-decoupled map $\widetilde{S}_{decouple}$ from Bicon map $\widetilde{C}$ via the RCA module and replaced the loss with $L_{decouple}$ and $L_{conmap}$, respectively. This experiment is denoted as Exp. 4 in Table \ref{ablation}. As seen, the $F_{ave}$ and $E_m$ values increased. This result shows that the RCA module effectively utilized the extracted edge features.

\textbf{The connectivity consistency loss.} To test the effectiveness of the connectivity consistency loss, we replaced $L_{conmap}$ with $0.8 \times L_{conmap}+0.2 \times L_{bimap}$; i.e., the $L_{con\_const}$ in Exp. 4. Thus, the total loss function for this experiment is $L_{con\_const}+L_{decouple}$. For this complete BiconNet model with backbone GCPANet (Exp. 5 in Table \ref{ablation}), all three metrics improved, which demonstrates the ability of the connectivity consistency loss to improve the results.

\begin{figure}[h]
\begin{center}
   \includegraphics[width=1\linewidth]{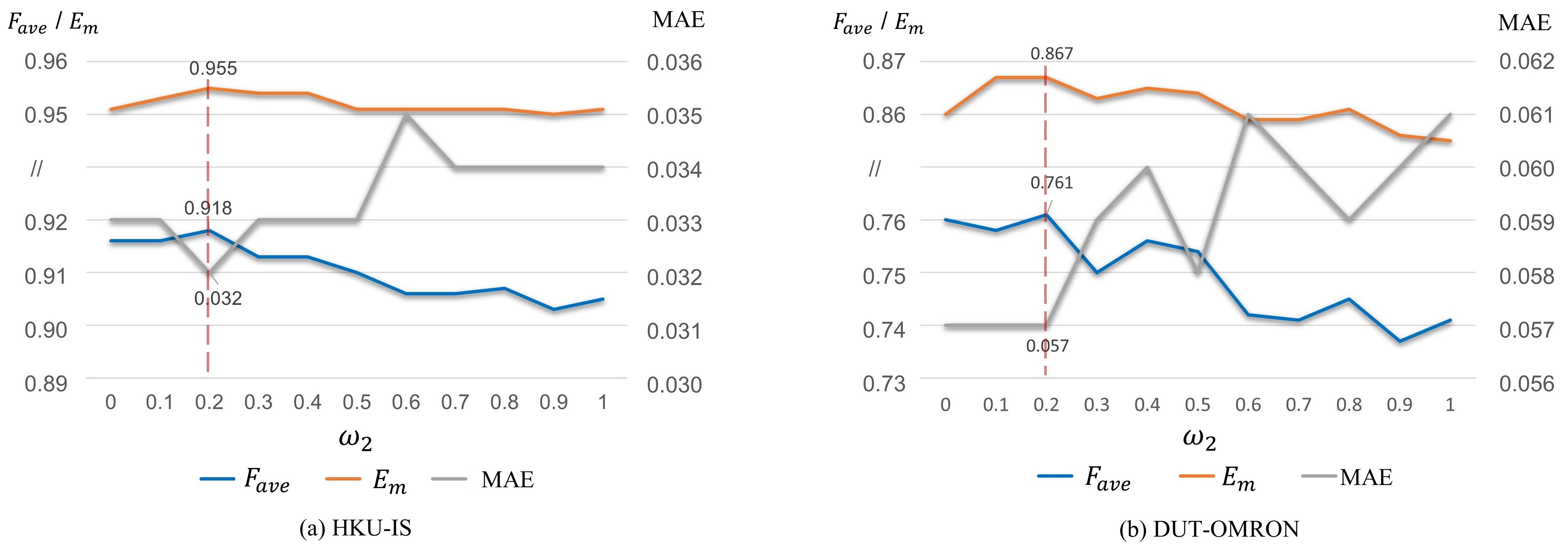}
\end{center}
\vspace{-10pt}
\caption{Training the network with different $\omega_1$  and $\omega_2$ on the (a) HKU-IS and (b) DUT-OMRON datasets. The x-axis represents the value for $\omega_2 \ (\omega_1=1- \omega_2)$. The best performance was achieved at $\omega_2=0.2$ (dashed red line).}
\label{weight}
\vspace{-5pt}
\end{figure}

\begin{figure}[h!]
\begin{center}
   \includegraphics[width=1\linewidth]{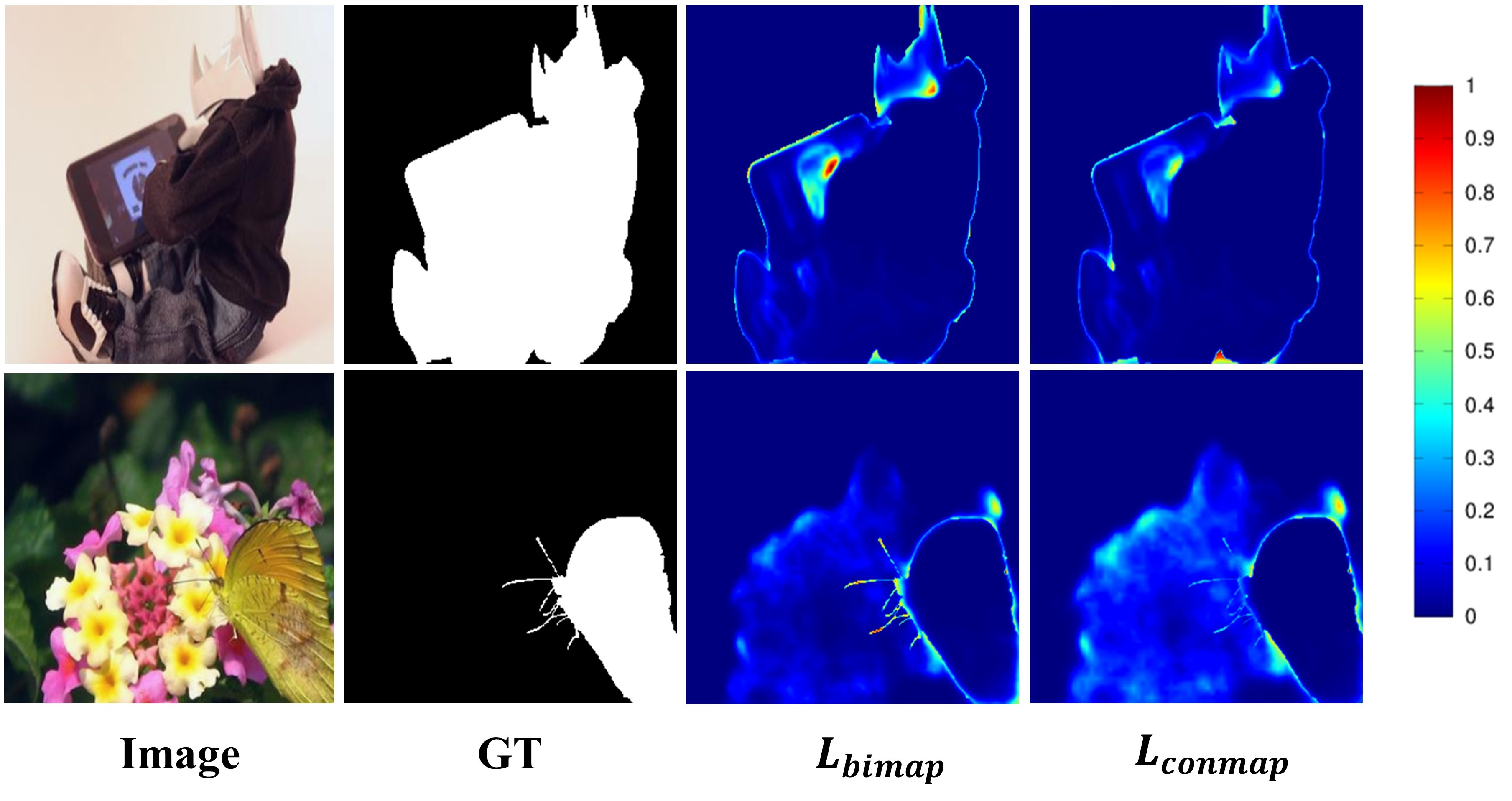}
\end{center}
\vspace{-10pt}
\caption{Comparison between $L_{bimap}$ and $L_{conmap}$. $L_{bimap}$ can generate much larger loss on hard pixels such as the edges of different objects while generating smaller loss on background pixels.}
\label{bimap_loss}
\vspace{-5pt}
\end{figure}

Additionally, to illustrate the different effects of $L_{conmap}$ and $L_{bimap}$ in $L_{con\_const}$, we conducted another set of experiments based on Exp. 5, using different weights for these two terms. The results are shown in Fig. \ref{weight}, where 10 experiments are plotted with $\omega_2$ as the x-axis ($\omega_1=1- \omega_2$). When we introduced $L_{bimap}$ and gradually increased its weight $\omega_2$ (from left to right), we observed that $F_{ave}$  and $E_m$ increased while MAE decreased at the beginning ($\omega_2 \leq 0.2$). Then, when $L_{bimap}$ had a larger weight, the overall performance decreased. The best performance was achieved at $\omega_1=0.8,\omega_2=0.2$. This result is consistent with our assumption that there is a tradeoff between edge enhancement and background dilution when using $L_{bimap}$. We also visualized the two loss terms $L_{bimap}$ and $L_{conmap}$ in Fig. \ref{bimap_loss} to further demonstrate this idea.

\subsection{Model Size and Testing Speed}
We list the model size and testing speed of our method in Table \ref{size}. To make fair comparisons, we reported the frame per second (FPS) processing speed with images of size $320 \times 320$ pixels for all listed models. Since we only changed the output fully-connected (FC) layers of the backbones, the increase in the parameters and testing time is directly related to the number of FC layers in the backbones. Thus, for those models with deep supervision mechanisms (multiple FC layers, such as GCPANet and EGNet), the increase is more than those using shallow supervisions. However, even for the deep supervised models, the increase of model size is still insignificant and the computational cost of our BiconNet is nearly identical to those of the baselines. Therefore, our method is efficient and can improve existing frameworks with neglectable increase in computational cost. 

\begin{table*}[h!]
\renewcommand\arraystretch{1.5}
\centering
\setlength{\tabcolsep}{1mm}{
\caption{Comparison of model size and testing speed between BiconNet and the corresponding baselines.}
\label{size}
\begin{tabular}{cccccccccccccccc} 

\hline
&\multicolumn{2}{c}{PoolNet}&\multicolumn{2}{c}{CPD-R}&\multicolumn{2}{c}{GCPANet}&\multicolumn{2}{c}{F3Net}&\multicolumn{2}{c}{EGNet}&\multicolumn{2}{c}{ITSD}&\multicolumn{2}{c}{MINet}\\
\cmidrule(lr){2-3}\cmidrule(lr){4-5}\cmidrule(lr){6-7}\cmidrule(lr){8-9}\cmidrule(lr){10-11}\cmidrule(lr){12-13}\cmidrule(lr){14-15}
&Base&Bicon&Base&Bicon&Base&Bicon&Base&Bicon&Base&Bicon&Base&Bicon&Base&Bicon\\
\hline
\makecell[c]{Number of\\Parameters(M)}&68.26&68.24&47.85&47.85&111.69&111.85&25.54&25.56&67.06&67.12&26.47&26.47&115.69&115.69\\ 
\hline
\makecell[c]{Testing Speed\\(FPS)}&49&49&55&53&38&34&64&63&60&53&47&44&45&43\\ 
\hline

\end{tabular}}
\end{table*}

\subsection{Compatibility Analysis}
In section 4.3, we showed that BiconNet is compatible with existing SOD frameworks in their \textit{entirety}. Here, we investigate the compatibility of BiconNet with \textit{individual modules} that have a similar function (i.e., enhancement of spatial coherence and edge modeling), such as inter-pixel consistency/edge-aware loss functions and CRF.

\subsubsection{With Pixel Relationship/Edge-Aware Loss Functions}
As illustrative examples to show the compatibility of BiconNet with the state-of-the-art loss functions, we considered two loss functions here: the Consistency enhanced Loss (CEL) \cite{MINet} (which can enhance the inter-pixel relationship) and Adaptive ConTour (ACT) \cite{ITSD} (which can improve the edge modeling). In each case, we compared the baselines and BiconNets with and without the loss functions (Table \ref{compatibility_1}). Again, the networks that included the BiconNet outperformed the baselines. We also note that the performance of ITSD didn’t significantly improve when added with ACT, while the combination of ACT and BiconNet had a more pronounced positive impact on performance.

\subsubsection{With CRF}
CRF is a widely used post-processing method that can enhance the inter-pixel relationship of the prediction and has been applied in SOD \cite{picanet,DSS,crf2}. Using GCPANet as our baseline, we added a fully connected CRF at the end of both GCPANet and GCPANet + Bicon for testing (Table \ref{compatibility_2}). The results of GCPANet + CRF show that $F_{ave}$ and MAE both improved while $E_m$ decreased compared to GCPANet.Similar results were observed in GCPANet + Bicon + CRF. Nonetheless, GCPANet + Bicon + CRF outperformed GCPANet + CRF, suggesting that BiconNet is compatible with CRF. 

When added to any model, CRF usually significantly increases the computational cost. However, the results show that our model (GCPANet + Bicon) can achieve comparable results with GCPANet + CRF (the 2nd and 3rd rows in Table \ref{compatibility_2}) without significantly compromising speed. 

\begin{table}[h!]
\renewcommand\arraystretch{1.2}
\centering
\setlength{\tabcolsep}{1.3mm}{
\small
\caption{Compatibility analysis with different loss functions.}
\label{compatibility_1}
\begin{tabular}{ccccccc} 
\hline
\multirow{2}*{Model}&\multicolumn{3}{c}{DUT-TE}&\multicolumn{3}{c}{DUT-OMRON}\\
\cmidrule(lr){2-4} \cmidrule(lr){5-7}
&F$_{ave}\uparrow$&MAE$\downarrow$&$E_m\uparrow$&F$_{ave}\uparrow$&MAE$\downarrow$&$E_m\uparrow$\\
\hline
ITSD \textbf{w/o} ACT&0.805&0.041&0.898&0.750&0.059&0.862\\ 
$+$Bicon&0.830&0.041&0.902&0.763&0.059&0.865\\ 
ITSD \textbf{w/} ACT&0.806	&0.041	&0.891&0.752&0.058&0.862\\ 
$+$Bicon&0.838	&0.038	&0.905	&0.774	&0.053	&0.874\\ 
\hline
MINet \textbf{w/o} CEL&0.801	&0.036	&0.901	&0.749	&0.053	&0.868\\ 
$+$Bicon&0.846	&0.037	&0.910	&0.766	&0.053	&0.870\\ 
MINet \textbf{w/} CEL&0.838	&0.035	&0.903	&0.762	&0.053	&0.870\\ 
$+$Bicon&0.856	&0.035	&0.915	&0.778	&0.051	&0.875\\ 
\hline
\end{tabular}}
\end{table}

\begin{table}[h!]
\renewcommand\arraystretch{1.2}
\centering
\setlength{\tabcolsep}{0.2mm}{
\small
\caption{Compatibility and testing speed analysis with CRF.}
\label{compatibility_2}
\begin{tabular}{ccccccc} 
\hline
\multirow{2}*{Model}&\multicolumn{3}{c}{DUT-TE}&\multicolumn{3}{c}{DUT-OMRON}\\
\cmidrule(lr){2-4} \cmidrule(lr){5-7}
&F$_{ave}\uparrow$&MAE$\downarrow$&$E_m\uparrow$&F$_{ave}\uparrow$&MAE$\downarrow$&$E_m\uparrow$\\
\hline
GCPANet&0.896	&0.032	&0.950	&0.743	&0.056	&0.856\\ 
GCPANet $+$ Bicon&0.918	&0.032	&0.954	&0.762	&0.055	&0.863\\ 
\hline
GCPANet $+$ CRF&0.920	&0.029	&0.947	&0.763	&0.053	&0.840\\ 
{GCPANet $+$ CRF $+$ Bicon}&0.928	&0.029	&0.950	&0.775	&0.051	&0.856\\ 
\hline
\end{tabular}}

\end{table}

\section{Conclusion}

In this study, we examined the spatial inconsistency and blurred edge issues of general salient object detection methods. To overcome these problems, we proposed a connectivity-based approach called BiconNet. We first showed that the connectivity mask is a more structure- and inter-pixel relation-aware label than a single-channel saliency mask. To utilize this informative label, we proposed a BV module to enhance the spatial consistency of the output and an RCA module to extract the edge features. Then, we trained the model with a novel Bicon loss. Extensive experiments demonstrated the advantages of our method over state-of-the-art algorithms. Finally, we demonstrated the efficiency of our model as it can improve existing SOD frameworks with a neglectable increase in computational cost.

Although this work demonstrated significant advances in the field of SOD, there are still properties of the connectivity mask worth exploiting in future work. For example, a weakness of our approach is that we only considered the inter-class connectivity for the single-class segmentation problem. When dealing with the multi-class segmentation task, our method is expected to further benefit from modeling the intra-class relationship between connectivity masks. We envision that our connectivity-based approach to the image segmentation problem can be adopted by us and others in a wide range of applications, including semantic segmentation, instance segmentation, and segmentation of medical images.

{\small
\bibliographystyle{ieee_fullname}
\bibliography{mybibfile}
}

\end{document}


\title{BiconNet: An Edge-preserved Connectivity-based Approach for Salient Object Detection --- Supplementary Material}

\author{First Author\\
Institution1\\
Institution1 address\\
{\tt\small firstauthor@i1.org}
\and
Second Author\\
Institution2\\
First line of institution2 address\\
{\tt\small secondauthor@i2.org}
}

\maketitle

\

This supplementary material provides complementary information about Bilateral Connectivity Network (BiconNet). 

\section{Analysis of the Model Size}
As illustrated in the paper, the numbers of parameters of proposed BiconNet are very closed to those of baselines since we only changed the last fully connection layers. This conclusion can be also viewed in Tab. \ref{size}. This means we can improve the baselines without generating too much extra computational costs.

\section{Experiment Setting Details}
As we illustrated in the paper, we adopted the same data pre-processing tricks and network settings as described the baselines' paper \cite{gcpa,poolnet,MINet,egnet,cpd,ITSD} for every baseline experiment. For the corresponding BiconNet version of the baseline, we mainly changed the starting learning rates of that of the baselines. The details of the parameter settings of all our experiments are shown in Tab. \ref{settings}. Note that we used the official codes for the baselines and strictly followed the instruction on their websites. All experiments were trained from scratch.
\

\begin{table}
\begin{center}
\setlength{\abovecaptionskip}{0.cm}
\caption{The model sizes of baselines and their corresponding BiconNet version.}
\label{size}
\includegraphics[width=0.98\linewidth]{{latex/size.png}}
\end{center}
\vspace{-9pt}
\end{table}
\section{The Effect of $L_{bimap}$}
\begin{figure*}[b]
\begin{center}
\setlength{\abovecaptionskip}{0.cm}
\includegraphics[width=0.95\linewidth]{{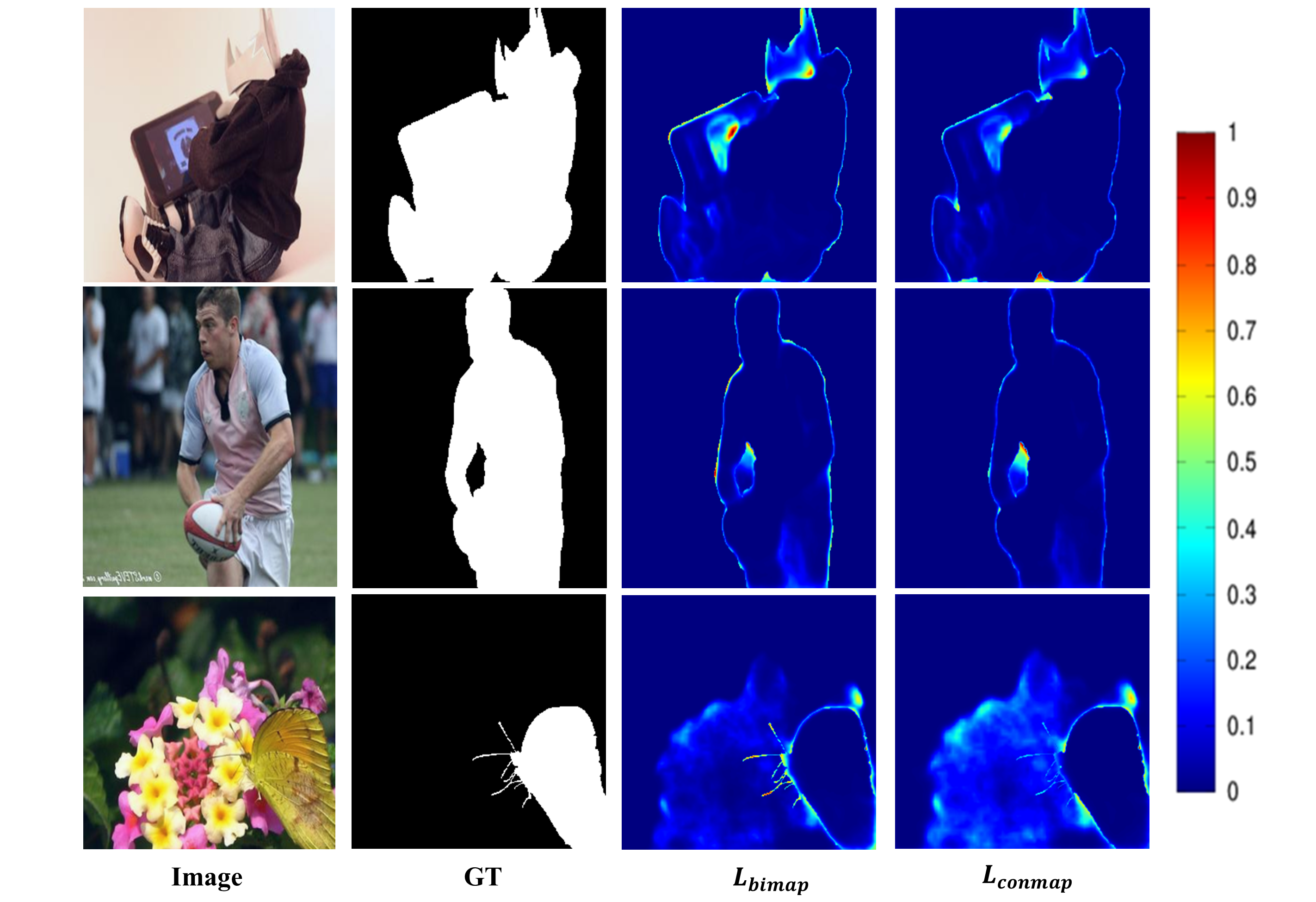}}
\caption{The visualization of $L_{bimap}$ and $L_{connmap}$. We can see that different from $L_{connmap}$ which puts global attention on the image, $L_{bimap}$ focuses more on the hard salient pixels, such as the edges of objects or the salient regions where the intensities have sharp changes, and generates much larger loss at these regions.}
\label{bimap}
\end{center}
\vspace{-9pt}
\end{figure*}

In this section, we will illustrate the effect of $L_{bimap}$ in the training. As mentioned in the paper, $L_{bimap}$ computes the multi-channel binary cross entropy loss based on the Bicon map $\widetilde C$. Different from Conn map $C$ which is the direct output of the network, Bicon map $\widetilde C$ is generated by multiplying the two unidirectional probability through bilateral voting. Therefore, the elements in $\widetilde C$ are exponentially correlated with those in $C$. For the hard salient pixel \cite{f3net}, such as an edge pixel, since it is likely to be predicted as a low probability value in $C$, it will generate an even smaller value in $\widetilde C$. This means the $L_{bimap}$ will generate much larger losses at these pixels as shown in Fig. \ref{bimap}. As a result, $L_{bimap}$ can encourage the network to pay more attention to the structural information and help the network to learn to handle the complex scenes.

\begin{table*}[b]
\begin{center}
\caption{The detailed network settings for BiconNets with different backbones.}
\label{settings}
\includegraphics[width=0.98\linewidth]{{latex/settings.png}}
\end{center}
\end{table*}

{\small
\bibliographystyle{ieee_fullname}
\bibliography{Bicon_library}
}